\documentclass[10pt,twocolumn,letterpaper]{article}

\usepackage[pagenumbers]{wacv}      % To produce the REVIEW version for the algorithms track
\usepackage{multirow}
\usepackage{xcolor,colortbl}
\usepackage{adjustbox}

\definecolor{wacvblue}{rgb}{0.21,0.49,0.74}
\usepackage[pagebackref,breaklinks,colorlinks,allcolors=wacvblue]{hyperref}
\usepackage[accsupp]{axessibility}
\usepackage[ruled]{algorithm2e} 
\usepackage{algorithmic}

\usepackage{placeins}

\newif\ifrevisionMode
\revisionModefalse % ← 黒で表示したいとき

\newcommand{\revision}[1]{%
  \ifrevisionMode
    \textcolor{red}{#1}%
  \else
    #1%
  \fi
}

\def\wacvPaperID{1768} % *** Enter the WACV Paper ID here
\def\confName{WACV}
\def\confYear{2026}

\title{ControlVP: Interactive Geometric Refinement of AI-Generated Images\\with Consistent Vanishing Points}

\author{Ryota Okumura \hspace{2em} Kaede Shiohara \hspace{2em} Toshihiko Yamasaki\\
The University of Tokyo\\
{\tt\small \{okumura, shiohara, yamasaki\}@cvm.t.u-tokyo.ac.jp}
}

\begin{document}
\twocolumn[{%
    \renewcommand\twocolumn[1][]{#1}%
    \maketitle
    
\begin{center}
  \includegraphics[width=\textwidth]{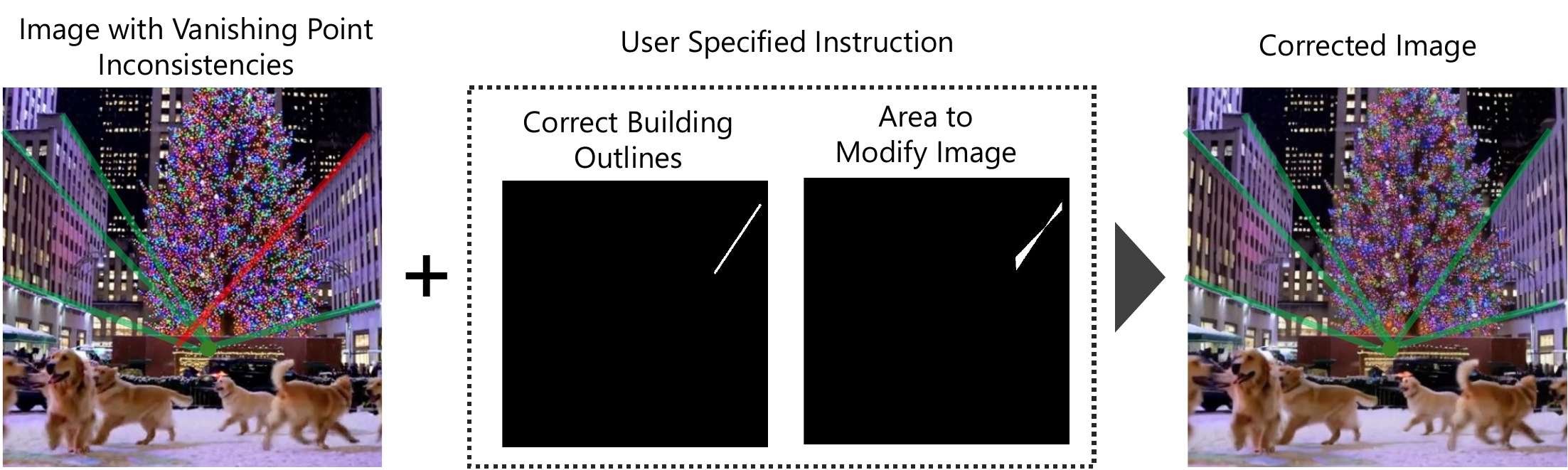}
  \captionof{figure}{Our method, ControlVP, corrects vanishing point (VP) inconsistencies in generated images. 
  In the left image, the green lines converge at a single VP while the red one does not, indicating a geometric inconsistency in the image.
  By incorporating user-provided building contours as conditions, our method transforms the inconsistent image into a geometrically coherent one where parallel lines properly converge at their respective VPs.
  Users specify desired / original building outlines through an interactive interface, allowing for precise geometric refinement.
  A mask is automatically generated from the areas between the original and desired outlines, limiting modifications to only the necessary regions.
  The initial image was adapted from Sora~\cite{videoworldsimulators2024}, a video generation model developed by OpenAI.}
  \label{fig:teaser}
\end{center}

}]
\begin{abstract}
Recent text-to-image models, such as Stable Diffusion, have achieved impressive visual quality, yet they often suffer from geometric inconsistencies that undermine the structural realism of generated scenes. One prominent issue is vanishing point inconsistency, where projections of parallel lines fail to converge correctly in 2D space. This leads to structurally implausible geometry that degrades spatial realism, especially in architectural scenes. We propose ControlVP, a user-guided framework for correcting vanishing point inconsistencies in generated images. Our approach extends a pre-trained diffusion model by incorporating structural guidance derived from building contours. We also introduce geometric constraints that explicitly encourage alignment between image edges and perspective cues. Our method enhances global geometric consistency while maintaining visual fidelity comparable to the baselines. This capability is particularly valuable for applications that require accurate spatial structure, such as image-to-3D reconstruction. The dataset and source code are available at \url{https://github.com/RyotaOkumura/ControlVP}.
\end{abstract}
    
\section{Introduction}
\label{sec:intro}

The emergence of models such as Stable Diffusion (SD)~\cite{Rombach_2022_CVPR} and Imagen~\cite{saharia2022photorealistic} 
has popularized the technology to generate high-quality images from text prompts. 
These models demonstrate a broad range of practical applications, 
from supporting artistic creation to content production and product design.  
Furthermore, higher-quality models such as SDXL~\cite{podell2023sdxl}, Kandinsky 3.0~\cite{arkhipkin2023kandinsky}, and PixArt~\cite{chen2024pixartalpha} continue to emerge, steadily improving the quality of generated images. 
However, even these state-of-the-art models still face challenges in preserving geometric consistency. 
This is evident in the form of vanishing points (VPs)~\cite{sarkar2024shadows}.

\begin{figure}[t]
  \centering
  \includegraphics[width=\linewidth]{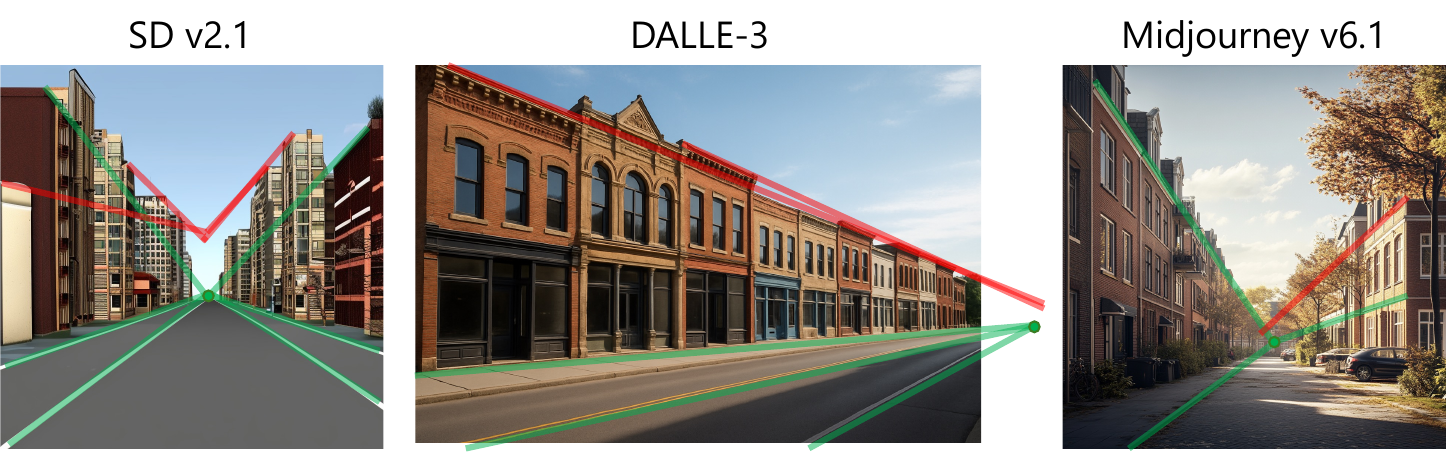}
  \caption{Examples of VP inconsistencies across different image generation models. 
  From left to right: Stable Diffusion v2.1~\cite{Rombach_2022_CVPR}, DALLE-3~\cite{dalle3} and Midjourney v6.1~\cite{midjourney}. 
  A set of parallel lines in 3D space (red and green) should form a single VP, but they converge at different points.}
  \label{fig:VPinconsistent_samples}
\end{figure}

A VP is a geometric concept describing the point in a 2D image where parallel lines in 3D space appear to meet.
Artists have understood the property since the Renaissance and used it as the foundation for perspective in paintings. 
This fundamental geometric constraint is not always maintained even in state-of-the-art image generation models, 
and parallel line groups frequently fail to form a single VP.
As illustrated in Figure~\ref{fig:VPinconsistent_samples}, 
this issue persists across leading models, including DALLE-3~\cite{dalle3}, Midjourney v6.1~\cite{midjourney}, and Stable Diffusion v2.1~\cite{Rombach_2022_CVPR}, 
where parallel building lines that should converge at common VPs instead converge at multiple inconsistent points.

This limitation restricts practical applications and diminishes the value of generated images.
For example, VPs are fundamental to human visual perception, serving as essential cues for understanding spatial depth and 3D structures.
VP inconsistency undermines the structural realism of generated scenes and limits their utility in fields requiring geometric precision, such as architecture design.
Moreover, VPs have proven valuable in various computer vision applications, including lane detection~\cite{yoo2017robust, wang2004lane,rasmussen2004grouping} and robotic navigation~\cite{lim2021avoiding, ji2015rgb}, where geometric inconsistencies prevent the practical use of AI-generated images in these downstream tasks.

To address this limitation, we propose ControlVP, a user-guided framework for correcting VP inconsistencies in generated images. 
Our approach extends a pre-trained diffusion model by incorporating structural guidance derived from building contours. 
We also introduce a novel ``VP loss'' that explicitly encourages alignment between image edges and perspective cues. 
During inference, users can interactively specify the desired position of VPs and correct building outlines through an intuitive interface. 
The combination of these components enables precise modification of geometrically inconsistent regions while preserving the visual quality of unchanged areas.

We show an example modification of a single video frame of Sora~\cite{videoworldsimulators2024} in Figure~\ref{fig:teaser}. 
The original image contains parallel building outlines (red and green lines) that converge at different points before modification. After our correction, all parallel line groups properly form a single VP. 
For quantitative evaluation, we introduce the first dataset of VP inconsistencies, which is carefully annotated with desired VP positions and building outlines.
We demonstrate that our approach outperforms the baselines in metrics of VP consistency, while maintaining visual quality.

In summary, our main contributions are as follows.
\begin{enumerate}
    \item We propose ControlVP, the first user-guided framework that enables users to interactively correct VP inconsistencies in AI-generated images by specifying desired VP positions and building outlines.
    \item We introduce a new VP loss function that explicitly constrains image edges to align with perspective cues, resulting in improved geometric consistency compared to baseline methods.
    \item We create the first dataset of VP inconsistencies with annotations for quantitative evaluation of geometric correctness in generated images.
    \item We demonstrate through extensive experiments that our approach achieves superior VP consistency while maintaining visual fidelity across diverse architectural scenes.
\end{enumerate}

\section{Related Works}
\label{sec:related_works}

\subsection{Utilization of Vanishing Points}
Vanishing points (VPs) have been significant in computer vision for many years. 
Early research in the 90s primarily focused on camera calibration, \ie, estimating internal parameters (focal length, principal point, and distortion) and external parameters (rotation, translation) from one or multiple images~\cite{caprile1990using, guillou2000using, cipolla1999camera}, without calibration targets (\eg, checkerboard).
Most recent work such as~\cite{zhao2024extrinsic} which proposes a method for external calibration of Light Detection and Ranging (LiDAR) and cameras using VPs, demonstrates that the use of VPs for calibration remains an active and evolving research topic.

VPs play a crucial role in understanding road scenes, particularly in the context of autonomous driving. 
In lane recognition~\cite{yoo2017robust, wang2004lane, rasmussen2004grouping}, estimating VPs helps predict the approximate direction of lanes, which is based on the characteristic that roads and lanes are often relatively parallel. 
This prediction enables robust lane recognition despite noise, shadows, or lighting changes.
Furthermore, Guo et al.~\cite{guo2024vanishing} leveraged VPs in video semantic segmentation, using VP cues to improve frame-to-frame correspondence.

In robotics, VPs contribute to robot localization and mapping. 
Their utility has been demonstrated particularly in Simultaneous Localization and Mapping (SLAM) technology~\cite{lim2021avoiding, ji2015rgb}. 
SLAM allows robots to estimate their position while creating a map of the surrounding environment in unknown settings. 
It is essential to real-world robotics applications, such as autonomous mobile robots and drones. 
In SLAM, detected VPs are leveraged for more stable position estimation and map creation.

\subsection{Conditional Image Generation}

\revision{Latent diffusion models~\cite{Rombach_2022_CVPR} have enabled high-quality image generation from text prompts~\cite{Rombach_2022_CVPR, podell2023sdxl,chen2024pixartalpha,vladimir-etal-2024-kandinsky}.
Beyond text conditioning, recent works have explored additional conditioning mechanisms to provide more precise control over the generation process.
ControlNet~\cite{zhang2023adding} enables spatial conditioning through various image-based inputs such as edge maps, depth maps, and human pose.
GLIGEN~\cite{li2023gligen} and InstanceDiffusion~\cite{wang2024instancediffusion} enable instance-level location specification by incorporating bounding boxes or other location inputs to control object placement.}

\revision{In parallel, diffusion-based inpainting methods have shown remarkable progress in reconstructing missing regions.
RePaint~\cite{lugmayr2022repaint} leverages denoising diffusion probabilistic models with a resampling strategy to maintain consistency between inpainted and unmasked regions.
LatentPaint~\cite{corneanu2024latentpaint} performs inpainting in the latent space of diffusion models, achieving both efficiency and high-quality results.}

\subsection{Vanishing Point Consistent Image Generation}\label{sec:existing_methods}

The only existing work by Rishi et al.~\cite{upadhyay2023enhancing} proposes a VP consistent image generation method.
They pointed out that while conventional diffusion models prioritize image quality and prompt conformity, they do not sufficiently consider geometric constraints, which results in line distortions and inaccurate perspective representations in generated images. 
Furthermore, they also mentioned the potential impact of these geometric inconsistencies on the performance of downstream tasks such as monocular depth estimation.

To address this issue, they fine-tuned a depth-conditioned latent diffusion model~\cite{Rombach_2022_CVPR} using an additional loss term considering VP consistency. 
The proposed ``Perspective Loss'' works as follows. First, the initial image is predicted using the noise estimated by the model. 
Then, for each annotated VP in the input image, the edge intensities along lines radiating from the VP are calculated. 
Finally, the loss compares these edge alignments between the predicted image and the actual initial image, ensuring that the predicted image maintains similar VP characteristics.

As a result, they achieved image generation with high geometric consistency in VPs by using depth maps as input conditions. 
Their fine-tuned model successfully aligns parallel line groups to form single VPs. 
On the other hand, this method has practical limitations since depth maps for desired output images are hardly available or difficult to obtain in real-world scenarios.

\section{Proposed Method}
\label{sec:proposed_method}

\begin{figure}[t]
    \centering
    \includegraphics[width=\linewidth]{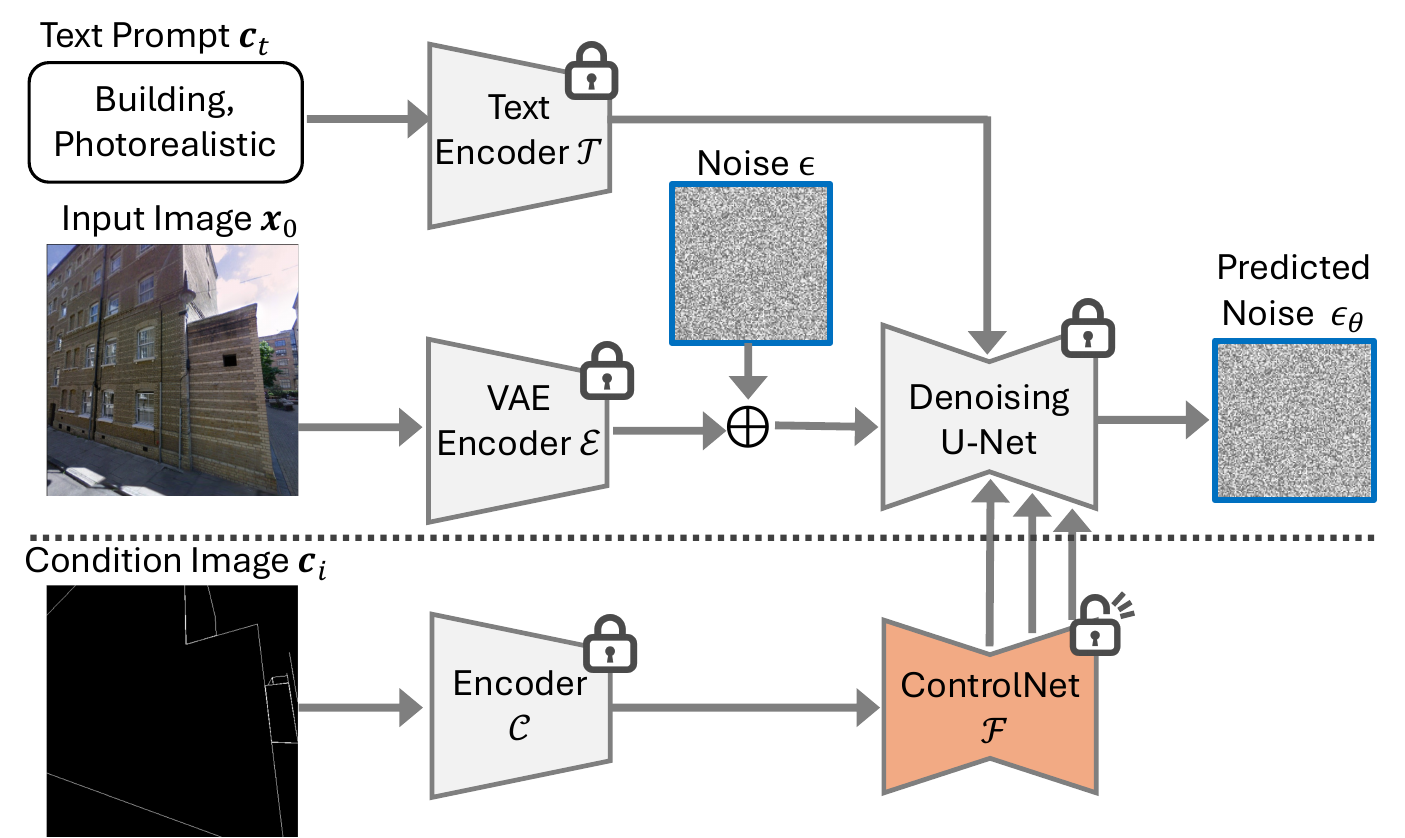}
    \caption{Overview of ControlVP. 
    The upper part shows the basic structure of LDM, where the VAE encoder compresses images into latent space, U-Net performs noise prediction, and the VAE decoder reconstructs the image. 
    The lower part illustrates the extended ControlNet architecture, which processes condition images (such as building outlines) and provides additional features to the U-Net, enabling geometrically consistent image generation.}
    \label{fig:controlnet}
\end{figure}

This section introduces ControlVP, a user-guided approach for correcting vanishing point (VP) inconsistencies in AI-generated images. 
We give the overview of our model in Figure~\ref{fig:controlnet}.
Our framework leverages a pre-trained Stable Diffusion v2.1 model enhanced with a ControlNet architecture to establish relationships between user-specified geometric instructions and the visual output. 
During training, we introduce a specialized Vanishing Point Loss that explicitly encourages alignment between image edges and perspective cues, to improve the model's ability to generate geometrically consistent images.

Users interact with our system through an intuitive graphical interface to correct VP inconsistencies during inference. 
The process starts with users identifying problematic regions and specifying both the desired VP positions and the correct building outlines. 
Users also indicate the original, incorrect outlines in the image. 
Based on these inputs, masks are created from the areas between the original and desired outlines, 
limiting modifications to only the necessary regions. 
We then apply an inpainting technique that selectively modifies these masked regions according to the provided geometric guidance 
while preserving the visual characteristics of unmodified areas.

\subsection{Preliminary - Latent Diffusion Model}

Latent Diffusion Models (LDMs) were first introduced alongside Stable Diffusion~\cite{Rombach_2022_CVPR}, 
which serves as the base model for our research. 
ControlNet~\cite{zhang2023adding} extends LDM's capability by enabling image conditioning.
Here, we explain the basic mathematical formulation of these models.

LDMs achieve high-quality image generation with relatively few inference steps by utilizing a Variational Autoencoder (VAE)~\cite{kingma2013auto} that compresses images into a lower-dimensional latent space where the diffusion process operates. 
The diffusion process adds noise to the original latent representation $\mathbf{z}_0$ to obtain the noisy latent $\mathbf{z}_t$ at time step $t$:
\begin{equation} \label{eq:diffusion_process}
        \mathbf{z}_t = \sqrt{\bar{\alpha}_{1:t}} \, \mathbf{z}_0 + \sqrt{1 - \bar{\alpha}_{1:t}} \, \epsilon,
\end{equation}
where $\epsilon \sim \mathcal{N}(0, \mathbf{I})$ is the noise sampled from the standard normal distribution, and $\bar{\alpha}_{1:t} = \prod_{i=1}^{t} \alpha_i$ is the cumulative product of the variance schedule.

ControlNet~\cite{zhang2023adding} extends this capability by allowing conditional image generation. 
It creates a trainable copy of the pretrained denoising U-Net~\cite{ronneberger2015u} that processes additional input conditions such as edges, poses, or, in our case, building outlines. 
The parameters are optimized so that the output of the U-Net $\epsilon_{\theta}$ matches the original image's actual noise $\epsilon$:
\begin{equation} \label{eq:controlnet_loss}
    \mathcal{L}_{cn} = \mathbb{E} \left[ \left\| \epsilon - \epsilon_{\theta}(\mathbf{z}_t, \mathcal{T}(\mathbf{c}_t), \mathcal{F}(\mathbf{c}_f), t) \right\|^2_2 \right], \\
\end{equation}
where $\mathbf{z}_t$ is the noisy latent, $\mathcal{T}(\mathbf{c}_t)$ is the encoded text prompt, and $\mathcal{F}(\mathbf{c}_f)$ is the processed condition image features. This enables the model to respect both semantic and structural guidance from text and visual inputs.

Classifier-free guidance (CFG)~\cite{ho2021classifierfree} is an inference-time technique that enhances the fidelity of input conditions by interpolating between conditional and unconditional noise predictions as follows:
\begin{equation} \label{eq:cfg}
    \hat{\epsilon}_{\theta} = (1 - \omega_{1})\epsilon_{\theta}^{uc} + \omega_{1} \epsilon_{\theta}^{c},
\end{equation}
where $\epsilon_{\theta}^{c}$ is the noise predicted with conditions, \eg, text prompts, $\epsilon_{\theta}^{uc}$ is the noise predicted without conditions, and $\omega_{1}$ is the strength of conditions.

\subsection{Adapting ControlNet for VP Consistency}
We treat the refinement of VP-inconsistent images as a spatially conditioned image generation task by user-specified correct outlines of buildings.
To this end, we adopt ControlNet for VP consistent image generation because of its powerful spatial controllability.
We train a ControlNet model as in Equation~\ref{eq:controlnet_loss}, where $\mathbf{z}_t$ and $\mathcal{T}(\mathbf{c}_t)$ are the latent of an outdoor scene photograph containing architectural structures and a consistent text prompt such as \textit{``modern buildings, high quality''}, respectively.
$\mathcal{F}(\mathbf{c}_f)$ represents a synthetic approximation of user-specified correct outline drawing images, created by randomly selecting building contour lines that converge toward a specific VP direction.

\begin{figure}[t]
    \centering
    \includegraphics[width=\linewidth]{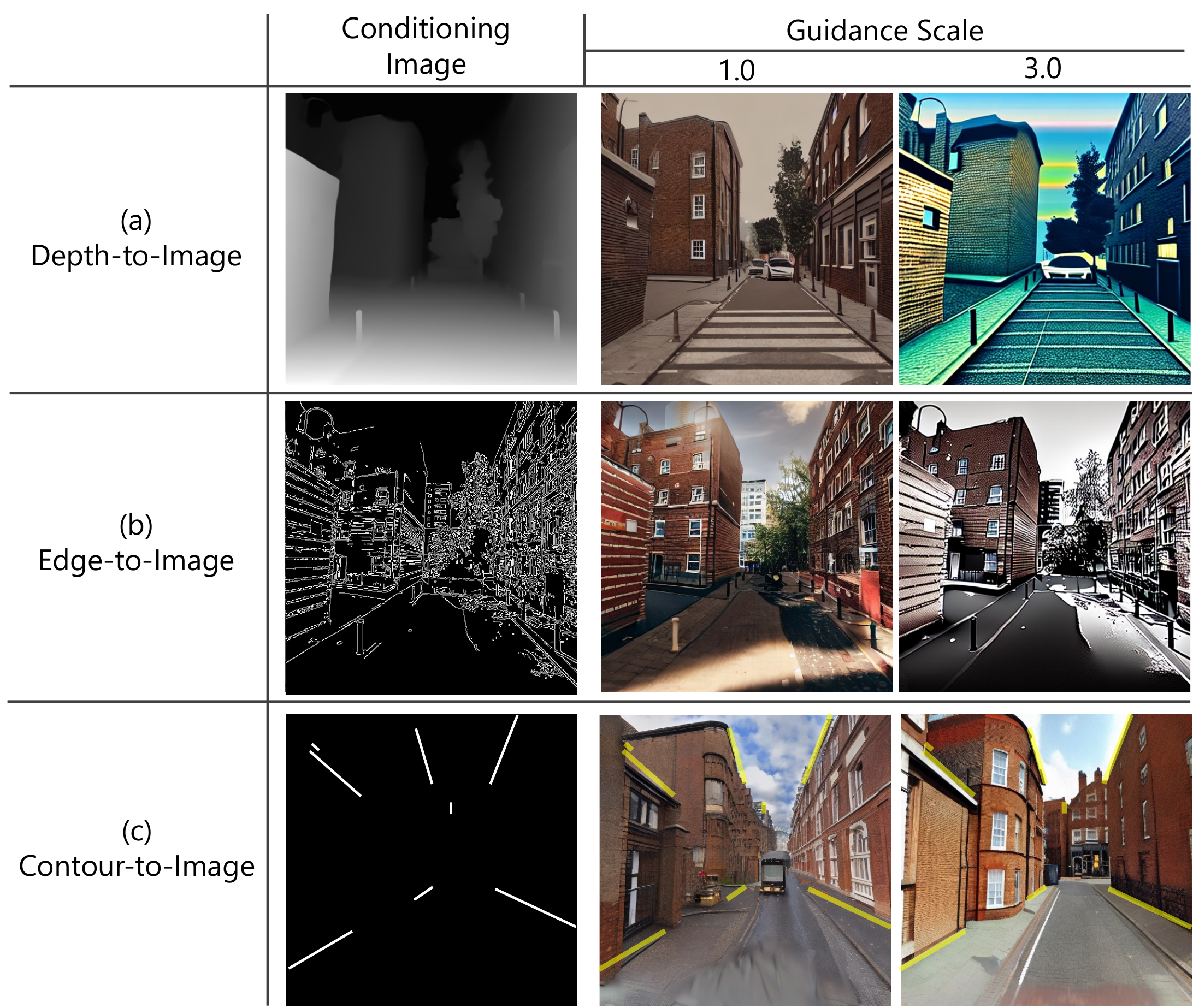}
    \caption{Classifier-free guidance (CFG) in different image translation tasks. 
    Images are generated by three different ControlNet conditions: (a) edge maps and (b) depth maps, and (c) contours towards VP. 
    The semitransparent yellow lines overlaid on the generated images of (c) indicate the given contour conditions.
    CFG harms the texture in conventional ControlNet tasks such as (a) and (b). 
    In contrast, CFG improves the conditioning fidelity while preserving the texture quality in our contour-to-image task (c).
    }
    \label{fig:cfg}
\end{figure}

\subsection{Classifier-Free Guidance for ControlVP}
We found it important to apply CFG not only to the text condition but also to the contour condition, unlike conventional ControlNet tasks where only the text CFG is applied.
During inference, at each denoising step $t$, we apply CFG to text and image conditions as follows:
\begin{equation} \label{eq:cfg_cond}
    \epsilon_{\theta}^{c} = (1 - \omega_{2})\epsilon_{\theta}^t(\mathbf{c}_t, \emptyset) + \omega_{2} \epsilon_{\theta}^t(\mathbf{c}_t, \mathbf{c}_f),
\end{equation}
\begin{equation} \label{eq:cfg_uncond}
    \epsilon_{\theta}^{uc} = (1 - \omega_{2})\epsilon_{\theta}^t( \emptyset, \emptyset) + \omega_{2} \epsilon_{\theta}^t( \emptyset, \mathbf{c}_f),
\end{equation}
where $\epsilon_{\theta}^t(\mathbf{c}_t, \mathbf{c}_f)$ denotes $\epsilon_{\theta}(\mathbf{z}_t, \mathcal{T}(\mathbf{c}_t), \mathcal{F}(\mathbf{c}_f), t)$ from Equation~\ref{eq:controlnet_loss}, and $\omega_{2}$ is the image guidance scale. We refer to image guidance scale when we mention ``guidance scale'' in the following sections.

As shown in Figure~\ref{fig:cfg}, in the conventional ControlNet tasks such as edge-to-image (a) and depth-to-image (b), CFG shows marginal structural differences between guidance scales, and suffers from notable texture degradation at a larger scale.
In contrast, our contour-to-image task (c) shows significant improvement with higher guidance scales, 
achieving better fidelity to the given condition at a scale of 3.0. 
This is because conventional ControlNet tasks typically focus on dense spatial conditioning, where models are trained such that the conditioning has a strong influence on the output. 
In contrast, in our task, the conditioning only affects a very small region of images, which results in a much weaker overall impact, allowing large guidance scales.

\subsection{Vanishing Point Loss} \label{sec:vanishing_point_loss}
To further improve the geometric consistency of the generated images, we introduce a new loss function called Vanishing Point Loss. 
This loss function compares edge intensities in the direction of VP between the generated and ground truth images.
While inspired by the Perspective Loss proposed in a prior work~\cite{upadhyay2023enhancing}, our Vanishing Point Loss introduces a fundamentally different design to better align with perceptual and geometric priorities. 

Firstly, it is required to decode an RGB image from the predicted latent $\hat{\mathbf{z}}_0$ to compute the edge of the predicted image.
A typical inference pipeline that requires dozens of inference steps from timestep $t$ to $0$ is not applicable here since it significantly increases computational costs for back-propagation.
To approximate this generation process efficiently, we adopt the 2-step denoising method~\cite{zhao2023diffswap}.
This method provides a better approximation than single-step denoising while remaining computationally feasible for training. 
Once the initial image is obtained, our loss computation process begins with applying Sobel filters to both the predicted initial image $\mathbf{\hat{x}}_0$ and the ground truth initial image $\mathbf{x}_0$ to obtain their edge directions and intensities.
We then analyze each VP in the ground truth image, calculating the sum of edge intensities for edges whose directions fall within a threshold angle of the VP direction.
The final loss is computed as the L2 difference between these sums, averaged across all VPs. % (Algorithm \ref{alg:vp_loss}).
Our overall loss function is a combination of the original ControlNet loss (Equation \ref{eq:controlnet_loss}) and our Vanishing Point loss:
\begin{equation}
    \mathcal{L}_{total} = \mathcal{L}_{cn} + \lambda \mathcal{L}_{vp},
\end{equation}
where $\lambda$ is a weighting factor that balances the contribution of geometric consistency against other aspects of image quality.
The complete formulation of $\mathcal{L}_{vp}$ and its algorithmic implementation are provided in Appendix \ref{sec:vp_loss_details}.

Importantly, the difference between our approach and the ``Perspective Loss'' proposed in the prior work~\cite{upadhyay2023enhancing} is that the latter directly uses the dot product between the edge direction and VP direction as weights for all edges. 
In contrast, our approach focuses on edges closely aligned with VP directions. 
This distinction is important because VP inconsistency manifests as minor angular errors that can significantly impact geometric perception. 
We place stronger constraints on edges that should align with VPs by applying a threshold angle and using sigmoid weighting that prioritizes edges within this threshold.
This targeted approach provides stronger geometric constraints while being differentiable and allowing for back-propagation.

\begin{figure}[t]
    \centering
    \includegraphics[width=\linewidth]{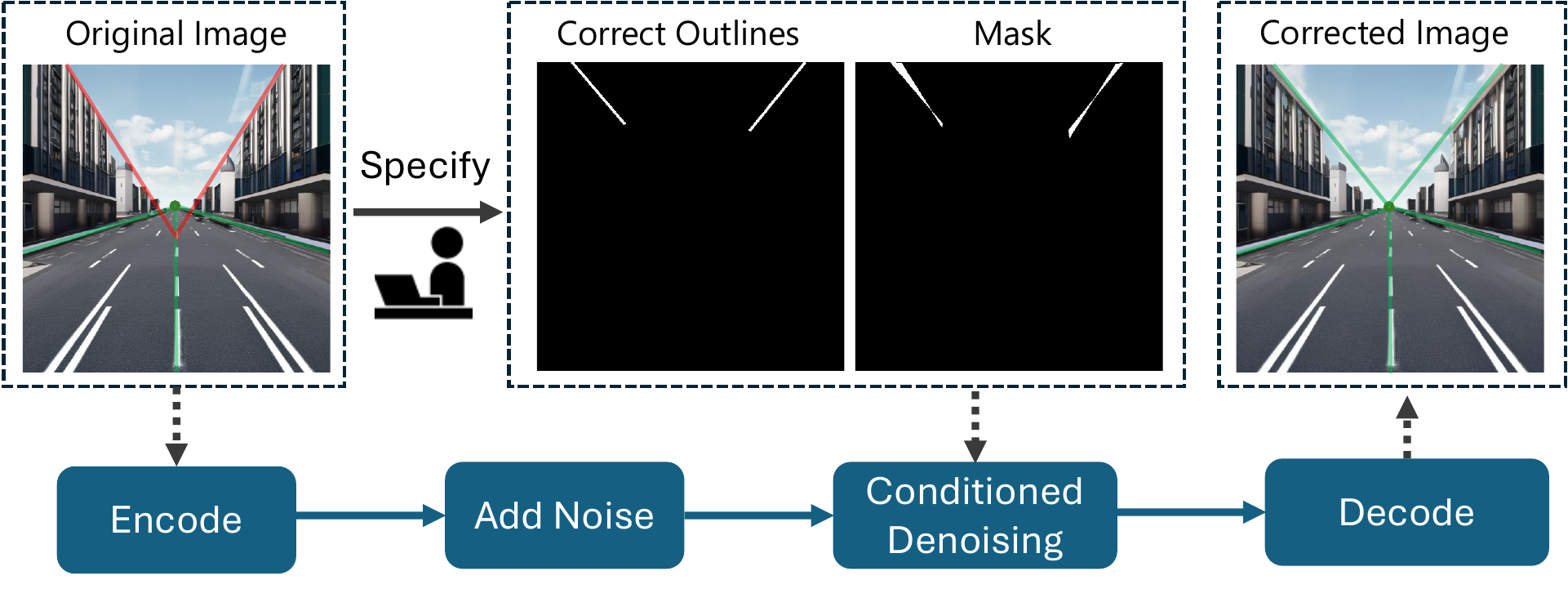}
    \caption{Inpainting process for VP correction. The process involves (a) mapping the input image to latent space, (b) adding noise to the latent representation, (c) performing denoising with predicted noise in the masked region while using true noise elsewhere, and (d) decoding the corrected latent to generate image with consistent VPs while preserving the unmasked regions.}
    \label{fig:inpainting}
\end{figure}

\subsection{Vanishing Point Correction Process} \label{sec:vanishing_point_correction_process}
We correct generated images during inference using an inpainting approach as shown in Figure~\ref{fig:inpainting}. 
This technique maps the input image to latent space and adds noise to create a noisy representation. 
The inpainting process uses our trained ControlNet model to predict noise for masked regions. 
For areas outside the mask, we directly use the ideal latent representation derived from the original image corresponding to each diffusion timestep. 
This approach ensures a seamless transition at mask boundaries, as both regions follow the same diffusion schedule but with different noise predictions. 
The resulting corrected image maintains continuity at the boundary while modifying only the targeted geometric structures inside the mask.
The user instruction for specifying target VPs and building outlines is performed through a graphical user interface (GUI) we developed.
See Appendix~\ref{sec:gui} for GUI details.

\section{Experiments}
\label{sec:experiments}

\subsection{Implementation Details}
\paragraph{Training Data.}
We selected HoliCity~\cite{zhou2020holicity} as our training dataset. 
This city-scale 3D dataset covers London suburbs and combines Google Street View panoramic images with accurately aligned CAD models. 
HoliCity provides 50,024 images with annotations, including normal maps, surface segmentation, depth maps, and VP positions.
Because HoliCity dataset lacks direct annotations of building outlines, we generate the annotations from the surface segmentation maps.
Concretely, we extracted contours and approximated them as polygons using the Douglas-Peucker algorithm~\cite{ramer1972,douglas1973}.
For each annotated VP in the dataset, we evaluated every polygon edge. We selected only edges with angle differences below our threshold $\theta$ relative to their corresponding VP direction. 
We then recorded the endpoint coordinates of these selected edges for each VP. 
This extraction process provided us with building outline data for all 50,024 images in the dataset.

\paragraph{Training Details.}
We used $N=4$ NVIDIA A100 (80 GB) GPUs fot training. 
The AdamW optimizer was used for optimization with a learning rate of $10^{-5}$. 
The batch size $B=2$ and gradient accumulation for $A=4$ steps resulted in a batch size of $B \times A \times N = 32$. 
We used gradient accumulation to maximize the effective batch size based on insights from the ControlNet implementation\footnote{\url{https://github.com/lllyasviel/ControlNet/blob/main/docs/train.md} (Accessed 2025-07-18)}, which suggests that training with a larger batch size is more effective than more training steps. 

\subsection{Setup}
\paragraph{Metrics.}
We adopted two metrics, angle accuracy (AA) and perceptual similarity distance (PSD), to measure the consistency with VP and overall image quality of the corrected images, respectively.
Our aim is to correct the VP inconsistency while preserving the original image characteristics, so we desire higher AA with smaller PSD.

AA is often used in VP detection tasks~\cite{lin2022deep, liu2021vapid, zhou2019neurvps}. 
This metric calculates the angle difference between direction vectors of parallel lines from detected VPs and ground-truth VPs in camera space and reports the ratio of VPs with differences below predefined thresholds.
We adapt AA using specified target VPs instead of ground truth VPs for our task. 
We employ the state-of-the-art VP detection method~\cite{lin2022deep} that does not rely on the Manhattan world hypothesis. 
When multiple VPs are detected, we select the one with the smallest angular error. 

We measure the PSD~\cite{zhang2018perceptual} between the original and corrected images. 
This metric aligns well with human visual judgment by capturing high-level semantic differences between images using deep neural network features.

\paragraph{Evaluation Dataset.}
\revision{We created three evaluation datasets from different text-to-image models: 250 images from SD v2.1~\cite{Rombach_2022_CVPR}, and 50 images each from FLUX.1-dev~\cite{flux2024} and PixArt-$\alpha$~\cite{chen2024pixartalpha}.}
These images primarily show buildings along straight roads as parallel line intentions are easier to identify.
We annotated each image with desired VP positions, correct building outlines, and original outlines in inconsistent regions with the same GUI employed for inference.
Additionally, Depth Anything V2~\cite{depth_anything_v2} provided metric depth maps necessary for comparison with the baseline~\cite{upadhyay2023enhancing}.

\subsection{Comparison with Previous Methods}
\paragraph{Compared Methods.}
We compare our approach with two baseline methods: (1) Depth-Persp, our adaptation of the existing method~\cite{upadhyay2023enhancing} for inpainting scenarios, where we apply their depth-conditioned model with perspective loss using depth maps from the original image as conditions; and (2) SD v2-Inpainting, which applies the standard Stable Diffusion v2 inpainting pipeline without geometric guidance. 
All methods use identical masks and inpainting processes to ensure fair comparison.

\begin{figure*}[t]
  \centering
  \includegraphics[width=1.0\textwidth]{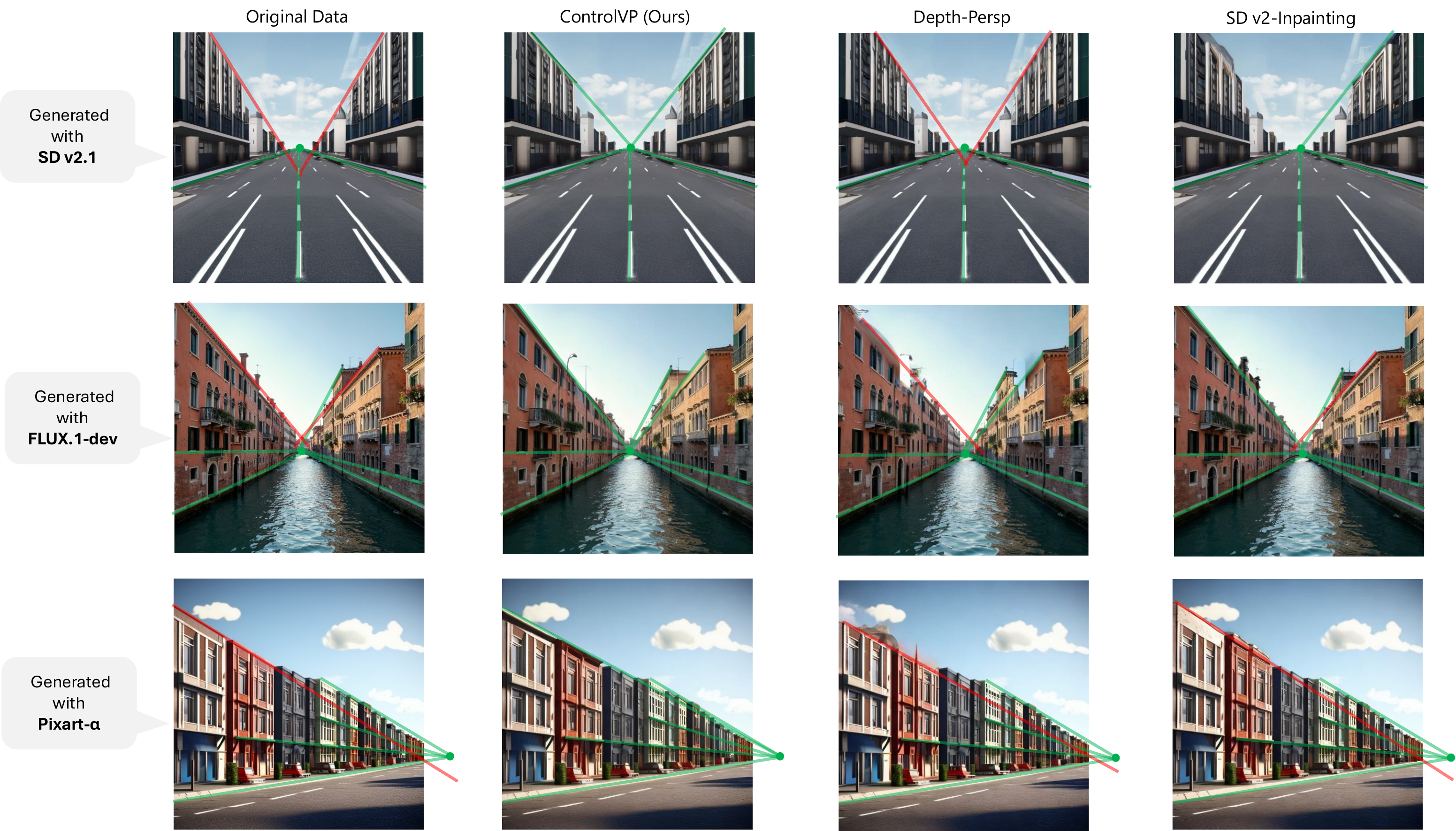}
  \caption{Qualitative comparison. 
  Red and green lines indicate parallel lines in 3D space. Outlines that match the specified vanishing point are overlayed in green, and inconsistent ones are shown in red.
  Depth-Persp uses depth maps as conditions, while SD v2.1 operates without additional geometric guidance. 
  ControlVP demonstrates superior geometric consistency, creating straighter building outlines that better align with the specified vanishing point direction.}
  \label{fig:qualitative_comparison}
\end{figure*}

\paragraph{Qualitative Result.}
Figure \ref{fig:qualitative_comparison} shows our qualitative comparison results.
We generated five images for each sample in the evaluation dataset and selected the best one that aligns optimally with the target VP.

Both Depth-Persp and SD v2-Inpainting often create edges that do not properly converge at the specified VP. 
Depth-Persp's outlines remain similar to the original distorted contours. This limitation comes from depth maps derived from already distorted original images. 
These distorted depth maps limit the model's correction ability despite perspective loss guidance. 
Our approach shows clear advantages through direct user instructions for image correction, 
especially for fixing already generated images.

\begin{table}[t]
    \centering
    \caption{\revision{Comparison with previous methods on three evaluation datasets.} The best and second best results among correction methods are highlighted in \textbf{bold} and \underline{underline}, respectively. 
    ControlVP achieves the superior angle accuracy across all thresholds and demonstrates consistent improvement over the original data.}
    \label{tab:comparison_previous_methods}
    \begin{adjustbox}{width=1.0\linewidth}
    \begin{tabular}{lcccc}
      \toprule
      Method & AA@3$^\circ$ & AA@5$^\circ$ & AA@10$^\circ$ & PSD \\
      \midrule
      \multicolumn{5}{l}{\textit{Dataset 1: Images from SD v2.1~\cite{Rombach_2022_CVPR} }} \\
      \midrule
      \textit{Original Data} & $0.610$ & $0.765$ & $0.908$ & - \\
      Depth-Persp & $0.642$ & $\underline{0.767}$ & $\underline{0.878}$ & $0.0283$ \\
      SD v2-Inpainting & $\underline{0.643}$ & $0.762$ & $0.877$ & $\mathbf{0.0258}$ \\
      ControlVP (Ours) & $\mathbf{0.731}$ & $\mathbf{0.826}$ & $\mathbf{0.910}$ & $\underline{0.0279}$ \\
      \midrule
      \multicolumn{5}{l}{\textit{Dataset 2: Images from FLUX.1-dev~\cite{flux2024}}} \\
      \midrule
      \textit{Original Data} & $0.700$ & $0.880$ & $0.980$ & - \\
      Depth-Persp & $\underline{0.706}$ & $\underline{0.868}$ & $\underline{0.984}$ & $0.0428$ \\
      SD v2-Inpainting & $0.698$ & $\underline{0.868}$ & $\mathbf{0.988}$ & $\mathbf{0.0347}$ \\
      ControlVP (Ours) & $\mathbf{0.772}$ & $\mathbf{0.876}$ & $\mathbf{0.988}$ & $\underline{0.0391}$ \\
      \midrule
      \multicolumn{5}{l}{\textit{Dataset 3: Images from PixArt-$\alpha$~\cite{chen2024pixartalpha}}} \\
      \midrule
      \textit{Original Data} & $0.760$ & $0.920$ & $0.940$ & - \\
      Depth-Persp & $\underline{0.776}$ & $0.858$ & $0.910$ & $0.0326$ \\
      SD v2-Inpainting & $0.762$ & $\underline{0.872}$ & $\underline{0.934}$ & $\mathbf{0.0280}$ \\
      ControlVP (Ours) & $\mathbf{0.886}$ & $\mathbf{0.942}$ & $\mathbf{0.972}$ & $\underline{0.0284}$ \\
      \bottomrule
    \end{tabular}
    \end{adjustbox}
\end{table}

\paragraph{Quantitative Result.}
Table \ref{tab:comparison_previous_methods} presents a quantitative assessment of our method compared to previous approaches and the baseline performance of the evaluation dataset itself.
\revision{Each method generated 10 images per sample, resulting in 2,500 images for the SD v2.1 dataset and 500 images each for the FLUX.1-dev and PixArt-$\alpha$ datasets.}

ControlVP consistently achieves the highest angle accuracy scores across all thresholds and datasets, outperforming both baseline methods.
Although the degree of improvement varies across source models, ControlVP demonstrates clear gains over the original images' geometric inconsistencies in all cases, whereas baseline methods often fail to improve or even worsen the original alignment.
The superior performance validates our design choices; our direct structural guidance by building outlines provides more effective control than indirect guidance by depth maps, and our specialized VP loss function successfully enforces geometric consistency during the generation process.

\subsection{Ablation Study and Analysis}

\begin{table}[t]
  \centering
  \caption{Effect of VP Loss. VP loss achieves the best performance across all angle accuracy metrics and the second best in perceptual distance compared to perspective loss and default loss.}
  \label{tab:ablation_vp_loss}
  \begin{adjustbox}{width=1.0\linewidth}
  \begin{tabular}{lcccc}
    \toprule
    Loss & AA@3$^\circ$ & AA@5$^\circ$ & AA@10$^\circ$ & PSD \\
    \midrule 
    $\mathcal{L}_{cn}$ & $\underline{0.678}$ & $0.788$ & $0.887$ & $\mathbf{0.0271}$ \\
    $\mathcal{L}_{cn} \& \mathcal{L}_{persp}$ & $0.677$ & $\underline{0.790}$ & $\underline{0.890}$ & $0.0298$ \\
    $\mathcal{L}_{cn} \& \mathcal{L}_{vp}$ (Ours) & $\mathbf{0.731}$ & $\mathbf{0.826}$ & $\mathbf{0.910}$ & $\underline{0.0279}$ \\
    \bottomrule
  \end{tabular}
  \end{adjustbox}
\end{table}

\begin{table}[t]
  \centering
  \caption{Effect of CFG. We compared three guidance scales (1.50, 3.00, 4.50) and four conditioning scales (1.25, 1.50, 1.75, 2.00) on SD v2.1 evaluation dataset. Default uses 1.0 for both scales.}
  \label{tab:ablation_cfg}
  \begin{adjustbox}{width=0.95\linewidth}
  \begin{tabular}{lcccc}
    \toprule
    Guidance & AA@3$^\circ$ & AA@5$^\circ$ & AA@10$^\circ$ & PSD \\
    \midrule 
    Default & $0.703$ & $0.812$ & $0.904$ & $\mathbf{0.0259}$ \\
    CFG 1.50 & $0.710$ & $0.808$ & $0.908$ & $\underline{0.0262}$ \\
    CFG 3.00 & $\mathbf{0.731}$ & $\mathbf{0.826}$ & $\mathbf{0.910}$ & $0.0279$ \\
    CFG 4.50 & $\underline{0.726}$ & $\underline{0.823}$ & $\underline{0.909}$ & $0.0302$ \\
    Cond 1.25 & $0.698$ & $0.812$ & $0.903$ & $0.0266$ \\
    Cond 1.50 & $0.701$ & $0.805$ & $0.898$ & $0.0274$ \\
    Cond 1.75 & $0.697$ & $0.806$ & $0.901$ & $0.0283$ \\
    Cond 2.00 & $0.696$ & $0.799$ & $0.897$ & $0.0294$ \\
    \bottomrule
  \end{tabular}
  \end{adjustbox}
\end{table}

\paragraph{Effect of VP Loss.}
To isolate the contribution of our VP loss, we compare three variants with identical architectures but different loss functions: (1) a model only with the default ControlNet loss $\mathcal{L}_{cn}$ (Equation \ref{eq:controlnet_loss}), (2) a model with $\mathcal{L}_{cn}$ and the perspective loss $\mathcal{L}_{persp}$ from prior work~\cite{upadhyay2023enhancing}, and (3) our model with $\mathcal{L}_{cn}$ and our VP loss~$\mathcal{L}_{vp}$.
Inference is performed with guidance scale $\omega = 3.0$ with SD v2.1 evaluation dataset.

Table \ref{tab:ablation_vp_loss} demonstrates the effectiveness of our proposed $\mathcal{L}_{vp}$ compared to $\mathcal{L}_{persp}$ and only using $\mathcal{L}_{cn}$.
Our VP loss consistently achieves the best performance across all angle accuracy metrics, while maintaining competitive perceptual distance.
The superior results validate our design choice of using threshold-based edge selection with sigmoid weighting, which provides stronger geometric constraints on edges aligned with VP directions compared to the uniform weighting approach used in $\mathcal{L}_{persp}$.

\paragraph{Effect of CFG.}
Table \ref{tab:ablation_cfg} evaluates the impact of CFG guidance scales (1.50-4.50) and conditioning scales (1.25-2.00) that amplify ControlNet outputs before U-Net integration, comparing against the default configuration.
The guidance scale results demonstrate a clear trade-off between geometric accuracy and perceptual quality. 
Higher guidance scales consistently improve angle accuracy metrics but increase perceptual distance from the original image. 
CFG with guidance scale 3.0 achieves optimal performance, delivering the best AA scores across all thresholds while maintaining reasonable image quality. 
Beyond 4.5, we observed qualitatively unnatural inpainting artifacts.
In contrast, conditioning scale adjustments show inconsistent improvements in geometric accuracy. 
Higher conditioning scales do not provide reliable performance gains, suggesting that CFG offers a more effective mechanism for enhancing geometric fidelity in VP correction tasks.

\paragraph{User Study.}
\revision{To assess the perceptual quality of our corrections, we conducted a user study with 20 participants.
We uniformly picked 36 images (12 from each evaluation dataset) with AA values greater than 3° to ensure that there are geometric inconsistencies to modify.
For each sample, participants selected the best result from five outputs generated by ControlVP,
and rated both alignment accuracy and naturalness on a 4-point scale. Details of the user study interface are shown in Appendix~\ref{sec:user_study_details}.}

\begin{figure}[t]
    \centering
    \includegraphics[width=1.0\linewidth]{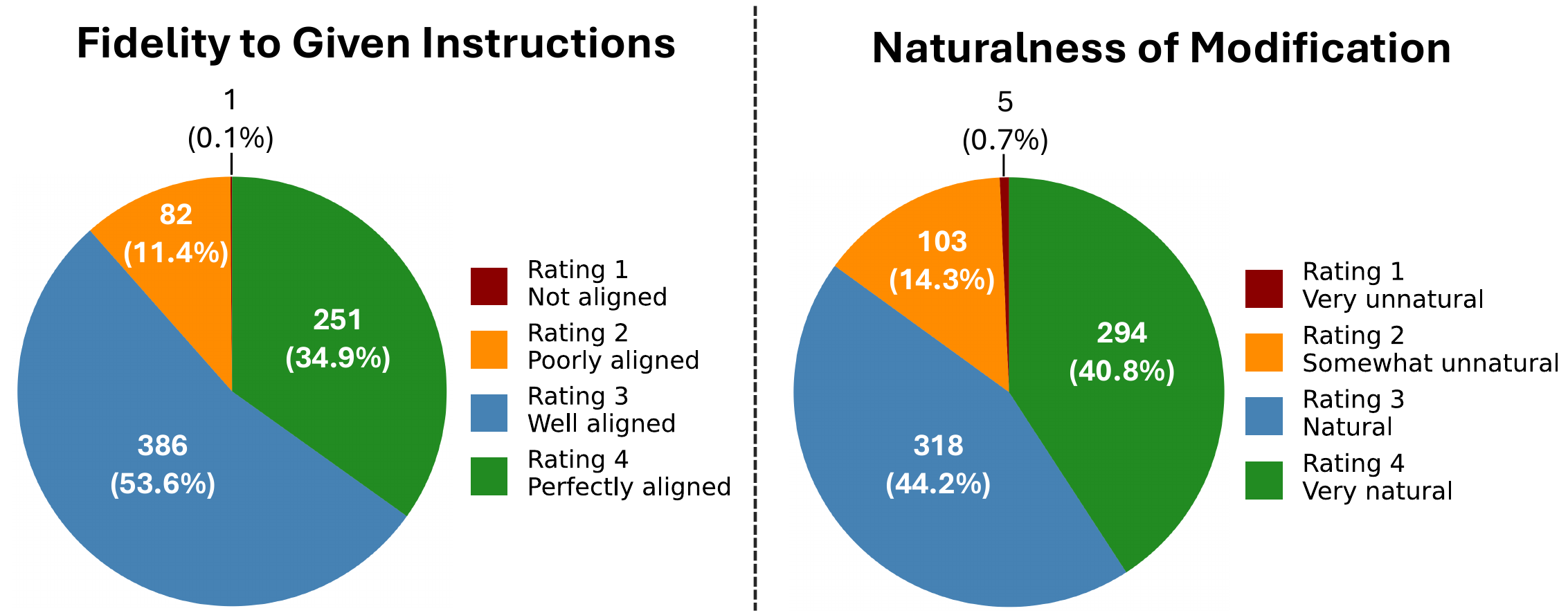}
    \caption{\revision{User study results comparing alignment accuracy and naturalness ratings.
    ControlVP achieves high scores for both geometric fidelity and visual quality, demonstrating effective VP correction while preserving image quality.}
    }
    \label{fig:user_study_results}
\end{figure}

\revision{Figure~\ref{fig:user_study_results} shows the distribution of user ratings for ControlVP's corrections.
For alignment fidelity, 88.5\% of responses rated the outputs as well-aligned (score 3) or perfectly aligned (score 4), with only 0.1\% indicating misalignment.
Similarly, for naturalness, 85.0\% of responses indicated natural (score 3) or very natural (score 4) results.
These results confirm that ControlVP successfully achieves accurate geometric corrections while preserving visual quality across diverse source models.}

\begin{figure}[t]
    \centering
      \includegraphics[width=1.0\linewidth]{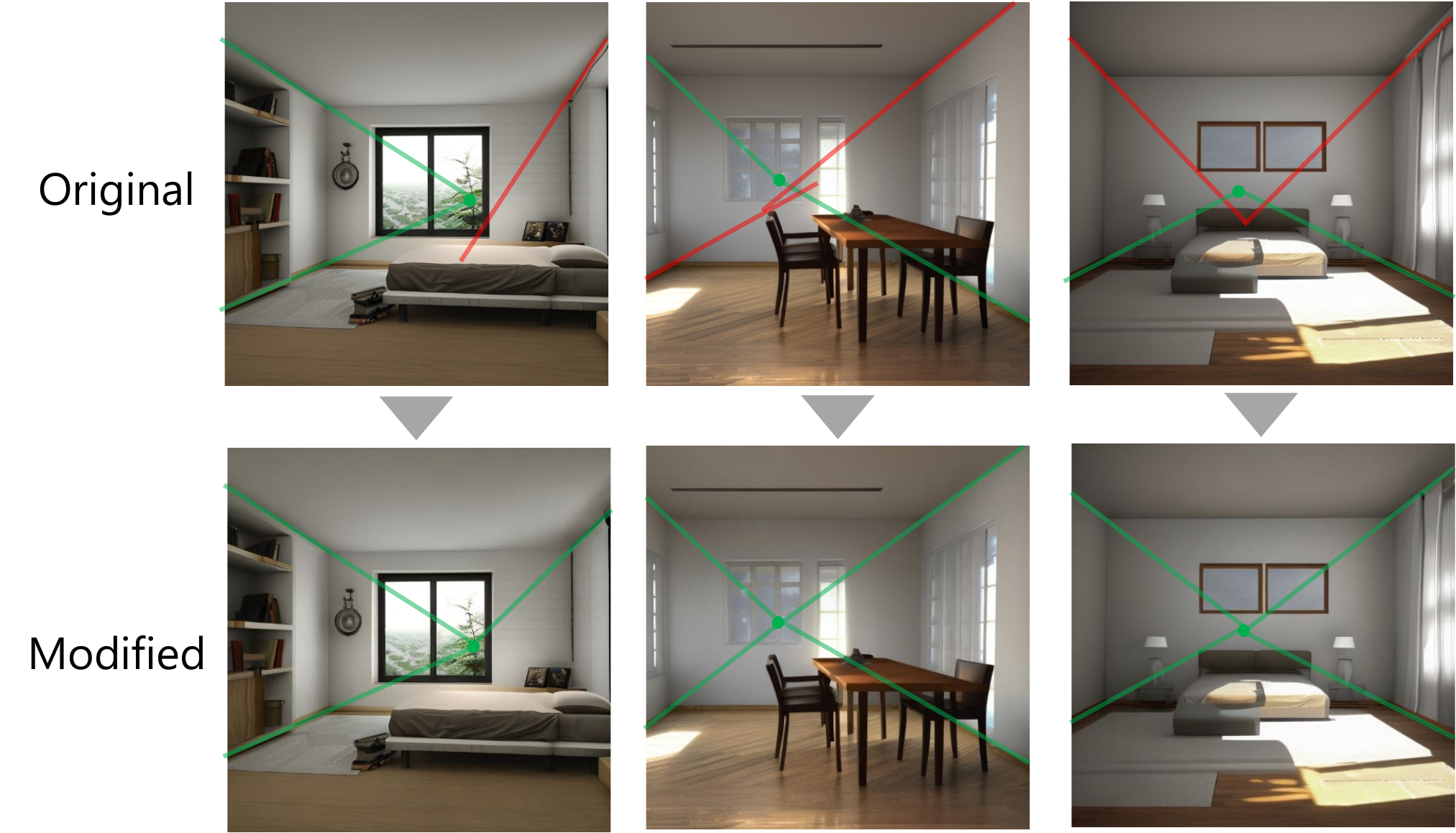}
    \caption{Indoor scene correction results. ControlVP successfully corrects vanishing point inconsistencies in indoor environments despite being trained only on outdoor architectural data. \revision{Original images are generated from SD v2.1.}}
    \label{fig:indoor_scene}
\end{figure}

\paragraph{Generalization to Indoor Scenes.}
Figure \ref{fig:indoor_scene} demonstrates the generalization capability of our ControlVP method to indoor environments despite being trained exclusively on outdoor architectural scenes from HoliCity. 
The results indicate that the learned geometric constraints for VP correction transfer effectively across different scene types, 
suggesting the broader applicability of our approach.

\section{Limitations}
\label{sec:limitations}
Our method demonstrates effectiveness when correcting building outlines to achieve VP consistency, representing a pioneering contribution in this unexplored domain. 
However, several limitations present opportunities for future work.
First, our approach targets buildings and cannot address inconsistencies in other scene elements (e.g., road markings or street lines). 
Users may need to adapt their correction strategy, potentially using misaligned road elements as reference points for building modifications.
\revision{Second, our workflow requires manual intervention, which can be a bottleneck when many images require correction.
Since parallelism in 3D space is fundamentally underdetermined from a single image, user specification of parallelism assumption is essential to resolve spatial discomfort.
We designed this process to be lightweight with automated VP detection and visual guides (see Appendix~\ref{sec:gui}).
Extending our approach to video is another promising direction, where maintaining consistent style corrections across frames would be key to achieving temporally coherent VP corrections.}

\section{Conclusion}
\label{conclusion}
We propose ControlVP, a novel method to correct VP inconsistencies in AI-generated images through user-guided refinement. 
Our approach extends a pre-trained diffusion model with a ControlNet architecture that processes building outline conditions. 
We also introduce a VP loss that explicitly encourages alignment between image edges and VP cues.
Our model offers practical advantages by requiring only visual instructions rather than additional depth information, 
making it accessible to users without specialized knowledge. 
Moreover, the intuitive GUI we developed enables users to specify VP positions and building outlines through a straightforward interface.
We create the first dataset of VP inconsistencies for comprehensive evaluation. 
The experimental results decisively demonstrate that our method outperforms baseline approaches in angle accuracy while maintaining comparable perceptual quality. 

\clearpage
{
    \small
    \bibliographystyle{ieeenat_fullname}
    \bibliography{bib/strings,bib/refs}
}
\clearpage
\appendix

\maketitlesupplementary

\noindent In this supplementary material, we provide the following:
\begin{enumerate}[label=\Alph*.]
    \item Detailed algorithm for VP loss computation.
    \item Graphical user interface for ControlVP.
    \item Evaluation dataset creation details.
    \item User study details.
    \item Additional qualitative results.
\end{enumerate}

\section{Implementation of Vanishing Point Loss} \label{sec:vp_loss_details}
The main paper introduces our Vanishing Point (VP) Loss and its conceptual design. 
Here, we provide the complete algorithmic implementation in detail and implementation considerations.

Our VP loss computation involves three key phases: edge detection, angular analysis, and weighted aggregation.
The algorithm operates on pixel-level edge information to enforce geometric consistency between predicted and ground truth images.

The algorithm takes three inputs: $\mathbf{x}_0$ is the clean ground truth image before noise addition, $\mathbf{\hat{x}}_0$ is the predicted clean image computed from the noisy latent and predicted noise, and $\{VP_i\}_{i=1}^N$ are the vanishing point coordinates annotated in the dataset for $\mathbf{x}_0$.

\begin{algorithm}[t]
    \caption{VP Loss Computation}
    \label{alg:vp_loss}
    \begin{algorithmic}[1]
    \STATE \textbf{Function} ComputeVPLoss($\mathbf{\hat{x}}_0$, $\mathbf{x}_0$, $\{VP_i\}_{i=1}^N$)
    \STATE \textbf{Input:} 
    \STATE \hspace{\algorithmicindent} $\mathbf{\hat{x}}_0$: Predicted image
    \STATE \hspace{\algorithmicindent} $\mathbf{x}_0$: Ground truth image
    \STATE \hspace{\algorithmicindent} $\{VP_i\}_{i=1}^N$: Set of vanishing point positions
    \STATE \textbf{Output:} $\mathcal{L}_{VP}$: Vanishing Point Loss
    
    \STATE \textbf{Step 1:} Compute edge using Sobel filter
    \STATE $(g_x^{pred}, g_y^{pred}) \gets \text{Sobel}(\mathbf{\hat{x}}_0)$
    \STATE $(g_x^{gt}, g_y^{gt}) \gets \text{Sobel}(\mathbf{x}_0)$
    \STATE $M^{pred} \gets \sqrt{(g_x^{pred})^2 + (g_y^{pred})^2}$
    \STATE $M^{gt} \gets \sqrt{(g_x^{gt})^2 + (g_y^{gt})^2}$
    
    \STATE \textbf{Step 2:} Compute edge direction scores
    \FOR{each pixel $(u,v)$}
        \STATE Compute edge direction vectors
        \STATE $\vec{d}^{pred} \gets (-g_y^{pred}, g_x^{pred})/M^{pred}$
        \STATE $\vec{d}^{gt} \gets (-g_y^{gt}, g_x^{gt})/M^{gt}$
        \FOR{each VP $VP_i$}
            \STATE Compute VP direction vector
            \STATE $\vec{v}_i \gets (VP_i^x - u, VP_i^y - v) / \|\vec{v}_i\|$
            \STATE Compute angle with edge direction
            \STATE $\theta^{pred}_i \gets \arccos(|\vec{d}^{pred} \cdot \vec{v}_i|)$
            \STATE $\theta^{gt}_i \gets \arccos(|\vec{d}^{gt} \cdot \vec{v}_i|)$
            \STATE Compute soft weight using sigmoid
            \STATE $w^{pred}_i \gets \sigma(\theta_{thresh} - \theta^{pred}_i)$
            \STATE $w^{gt}_i \gets \sigma(\theta_{thresh} - \theta^{gt}_i)$
            \STATE $\mathcal{S}^{pred}_i \gets \mathcal{S}^{pred}_i + M^{pred}w^{pred}_i$
            \STATE $\mathcal{S}^{gt}_i \gets \mathcal{S}^{gt}_i + M^{gt}w^{gt}_i$
        \ENDFOR
    \ENDFOR
    
    \STATE \textbf{Step 3:} Compute final loss
    \STATE $\mathcal{L}_{vp} \gets \frac{1}{N}\|\mathcal{S}^{gt} - \mathcal{S}^{pred}\|_2^2$
    \RETURN $\mathcal{L}_{vp}$
    \end{algorithmic}
\end{algorithm}

\textbf{Edge Detection:} We employ Sobel operators to extract both magnitude and direction information from images. 

\textbf{Angular Analysis:} For each pixel, we compute the angle between the edge direction and each vanishing point direction. Unlike previous approaches that apply uniform weighting across all edges, our method focuses on edges that are closely aligned with VP directions through threshold-based filtering.

\textbf{Weighted Aggregation:} We introduce a sigmoid weighting mechanism $\sigma(\theta_{thresh} - \theta_i)$ that provides differentiable weighting while maintaining selectivity for relevant edges. This differentiable formulation enables end-to-end training through backpropagation.

\section{Graphical User Interface for ControlVP} \label{sec:gui}

\begin{figure}[t]
    \centering
    \includegraphics[width=\linewidth]{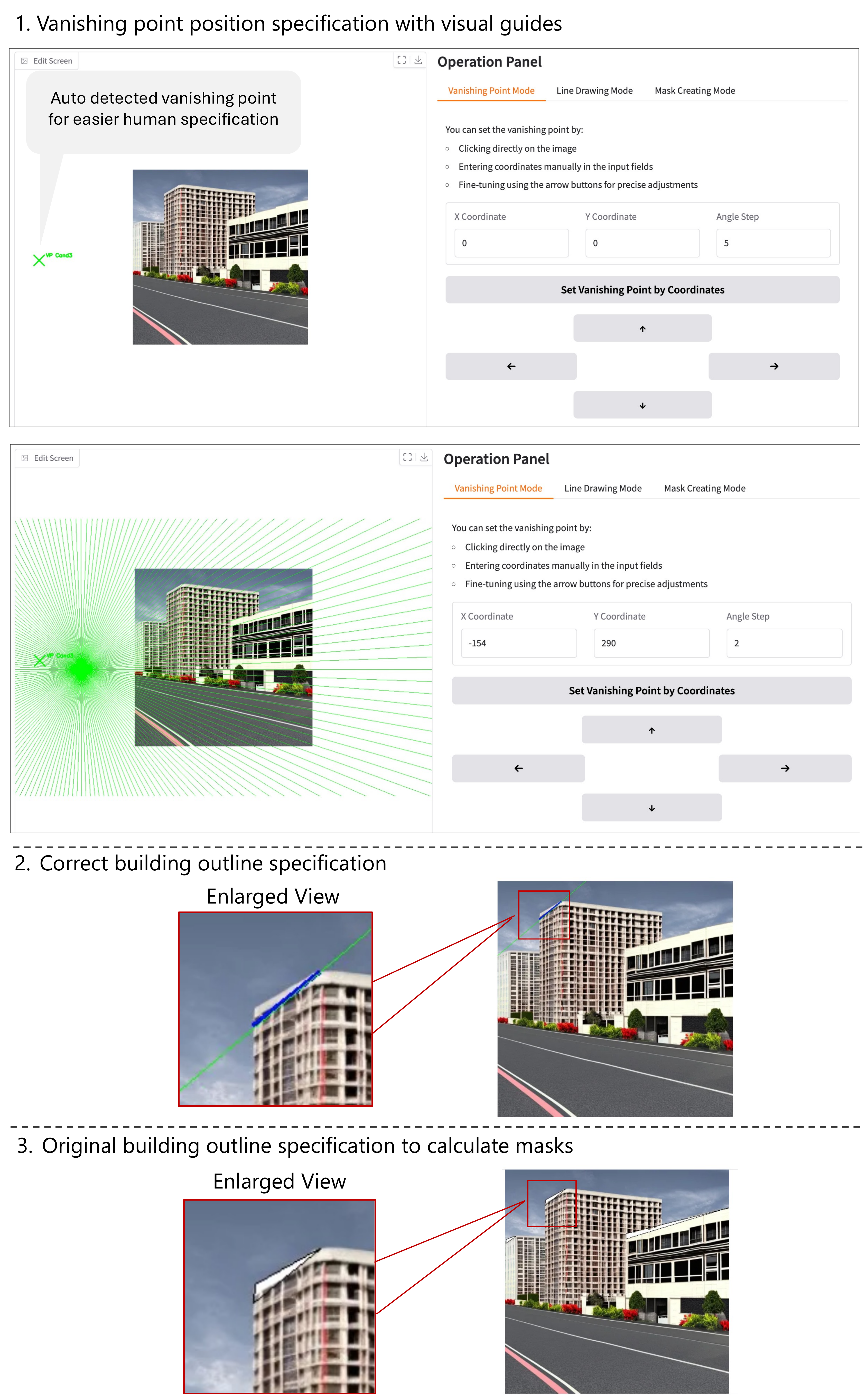}
    \caption{\revision{Graphical user interface for vanishing point correction.
    (1) Users select a VP position with the assistance of auto-detected VP suggestions displayed, simplifying the specification process.
    The system overlays auxiliary lines from the selected VP as visual guides.
    (2) Users mark the correct building outlines that should align with the VP.
    (3) Users specify the original misaligned outlines to define the mask regions for correction.}
    }
    \label{fig:GUI}
\end{figure}

\revision{
Please watch the supplementary video for a demonstration of our interface.
We developed an intuitive graphical user interface (GUI) shown in Figure \ref{fig:GUI} to make the correction process accessible to users without technical expertise.
The correction process is straightforward: (1) a single tap to select the VP position, (2) two taps to mark the endpoints of the correct building outline that should align with the VP, and (3) two taps to mark the endpoints of the original misaligned outline.
With a minimum of five taps per inconsistent region, users can specify the correct outlines and define the modification area as the region between the desired and original outlines.
}

\revision{
The system streamlines the correction workflow through automated assistance.
We employ an off-the-shelf VP detection model~\cite{lin2022deep} to automatically detect and display potential VP candidates, allowing users to easily select or refine the desired VP position based on these suggestions.
After VP selection, auxiliary lines are overlaid from the selected VP, serving as visual guides that help users easily identify where building outlines deviate from proper perspective.
For each identified inconsistency, users mark the desired and original outlines, and the system automatically generates a mask from the area between these outlines.
This targeted masking approach is crucial to the precision of our method, as it restricts modifications exclusively to geometrically inconsistent regions, preserving the original image's visual characteristics elsewhere.
}

\section{Evaluation Dataset Creation Details}

\revision{We generated evaluation images using the following prompts across three text-to-image models:}

\revision{\begin{enumerate}
    \item ``Buildings on both sides of the road, high quality, photorealistic''
    \item ``Row of buildings alongside a straight road, high quality, photorealistic''
    \item ``Row of buildings, high quality, photorealistic''
\end{enumerate}}

\revision{Each prompt was used to generate multiple images with SD v2.1~\cite{Rombach_2022_CVPR}, FLUX.1-dev~\cite{flux2024}, and PixArt-$\alpha$~\cite{chen2024pixartalpha}, resulting in three distinct evaluation datasets.
For each selected image, we manually annotated the target vanishing point, correct building outlines, and modification areas using the GUI described in Section~\ref{sec:gui}, and selected images with VP errors.
Depth maps for each image were generated using Depth Anything V2~\cite{depth_anything_v2} for use with Depth-Persp.
This resulted in 250 samples for SD v2.1, and 50 samples each for FLUX.1-dev and PixArt-$\alpha$.}

\section{User Study Details} \label{sec:user_study_details}

\begin{figure}[t]
    \centering
    \includegraphics[width=\linewidth]{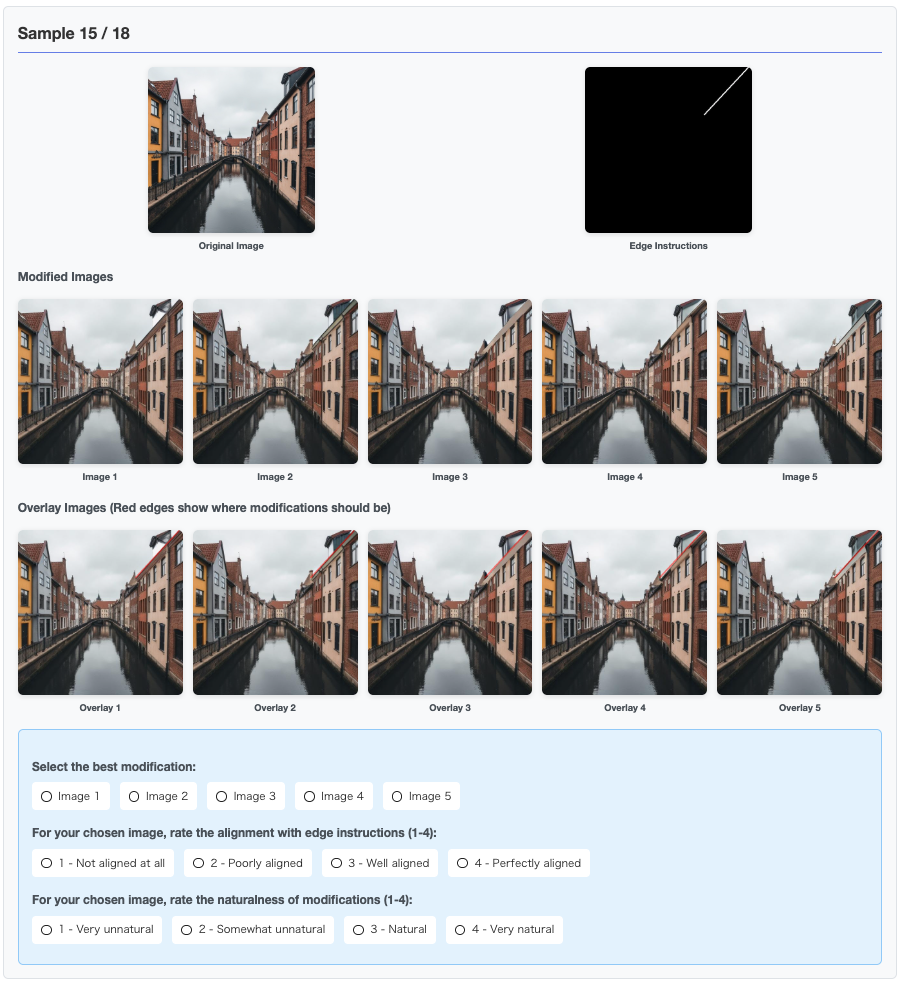}
    \caption{\revision{User study interface.
    Participants evaluate five corrected images for each sample, selecting the one that best follows the edge instructions while maintaining natural appearance.
    The interface displays the original image, edge instructions showing desired corrections, and five generated variations with overlay visualizations highlighting modification areas.}
    }
    \label{fig:user_study}
\end{figure}

\revision{We conducted a user study to evaluate the perceptual quality and geometric fidelity of our correction results.
Figure~\ref{fig:user_study} shows our user study interface, which presents participants with the original image alongside five consecutive correction results generated by our method.
To help participants understand the intended modifications, we provide edge instructions and overlay visualizations showing where corrections should occur. This user study is approved by the IRB of Graduate School of Information Science and Technology, The University of Tokyo (UT-IST-RE-250508\_6).}

\paragraph{Study Design.}
\revision{
In practical image generation workflows, users typically select the best images from multiple generated candidates. 
Therefore, it is important to evaluate the best rather than the average performance across multiple generations.
For each sample, we ask participants to:  (1) select the image that best follows the provided edge instructions, and (2) rate both the alignment accuracy (1-4 scale: not aligned to perfectly aligned) and naturalness of modifications (1-4 scale: very unnatural to very natural) for their chosen image.

We randomly sampled 12 images from each of our three evaluation datasets (SD v2.1, FLUX.1-dev, and PixArt-$\alpha$), selecting only images with AA greater than 3° to ensure meaningful geometric inconsistencies requiring correction.
Twenty participants completed the study, providing 720 total evaluations (20 participants × 12 samples × 3 models).}

\section{More Qualitative Results}
\begin{figure*}[t]
    \centering
    \includegraphics[width=0.97\linewidth]{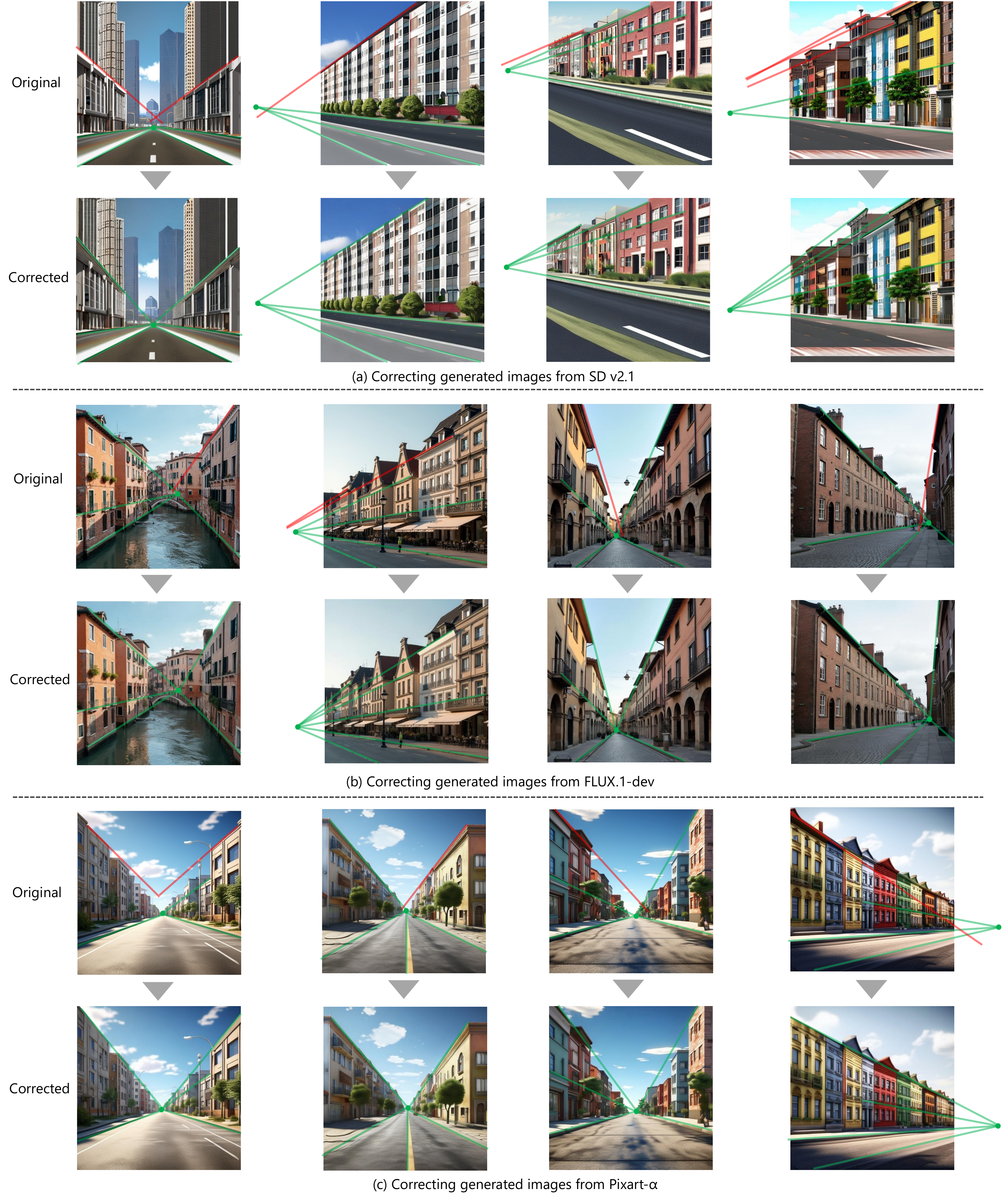}
    \caption{\revision{Additional qualitative results of ControlVP on images generated from three different text-to-image models.
    (a) Results on SD v2.1 generated images, (b) Results on FLUX.1-dev generated images, and (c) Results on PixArt-$\alpha$ generated images.
    For each example, we show the original image with VP inconsistencies (top) and the corrected result using ControlVP (bottom).
    Red lines indicate misaligned building outlines, while green lines show properly aligned outlines after correction.
    ControlVP successfully corrects geometric inconsistencies across diverse architectural styles and perspectives while maintaining visual quality.}
    }
    \label{fig:more_qualitative_example}
\end{figure*}

\revision{Figure~\ref{fig:more_qualitative_example} presents additional qualitative results of ControlVP on representative samples from our three evaluation datasets.
The examples include various architectural scenes with different styles and perspectives generated by SD v2.1, FLUX.1-dev, and PixArt-$\alpha$.
For each image pair, ControlVP successfully corrects the vanishing point inconsistencies while preserving the original visual characteristics.
The corrected images show proper alignment of building outlines toward the specified vanishing point, demonstrating consistent performance across different source models and scene types.}

\end{document}